\documentclass[10pt,twocolumn,letterpaper]{article}

\usepackage[pagenumbers]{cvpr} 
\usepackage{multirow}
\usepackage{soul}
\usepackage{textcomp,booktabs}
\usepackage{amssymb}
\usepackage{pifont}
\usepackage{makecell}

\usepackage{booktabs}
\usepackage{amsmath}
\usepackage{algorithm}        
\usepackage{algorithmic}      

\usepackage{amsfonts}
\usepackage{dblfloatfix}
\usepackage{graphicx}
\usepackage{textcomp}
\usepackage{caption}
\usepackage{subcaption}
\usepackage{atbegshi}
\usepackage{xstring}
\usepackage{flushend}
\usepackage[normalem]{ulem}
\usepackage{tikz}
\usepackage{multirow}
\usepackage{color}
\definecolor{forestgreen}{RGB}{0,139,69}
\usepackage{xcolor}
\definecolor{citecolor}{HTML}{0071bc}
\definecolor{SeaGreen4}{RGB}{0,205,102} 
\definecolor{SlateBlue}{RGB}{106,90,205} 
\definecolor{DarkRed}{RGB}{178,34,34} 
\usepackage[colorlinks, linkcolor=red,  anchorcolor=DarkRed, citecolor=SeaGreen4]{hyperref} 
\usepackage{makecell}
\usepackage{mathrsfs}
\usepackage{soul}
\usepackage{colortbl}
\usepackage{pgfplotstable} 
\usepackage{pgfplots}
\pgfplotsset{compat=newest} 
\usepackage{pgffor} 
\usepackage{filecontents}
\usepackage{enumitem}
\usepackage{pifont} 
\usepackage{bm}

\definecolor{cvprblue}{rgb}{0.21,0.49,0.74}

\def\paperID{*****} 
\def\confName{CVPR}
\def\confYear{2025}

\title{Long-Term Visual Object Tracking with Event Cameras: An Associative Memory Augmented Tracker and A Benchmark Dataset}

\author{Xiao Wang$^{1}$, Xufeng Lou$^{1}$, Shiao Wang$^{1}$, Ju Huang$^{1}$, Lan Chen$^{2}$, Bo Jiang$^{1}$ \\ 
    ${^1}${School of Computer Science and Technology, Anhui University, Hefei 230601, China} \\ 
    ${^1}${School of Electronic and Information Engineering, Anhui University, Hefei 230601, China.} \\ 
    \textit{louxufeng@stu.ahu.edu.cn}, \textit{wsa1943230570@126.com}, \textit{huangju991011@163.com}, \\ 
    \textit{\{xiaowang, chenlan, jiangbo\}@ahu.edu.cn}    
}

\begin{document}
\maketitle

\begin{abstract}
Existing event stream based trackers undergo evaluation on short-term tracking datasets, however, the tracking of real-world scenarios involves long-term tracking, and the performance of existing tracking algorithms in these scenarios remains unclear. In this paper, we first propose a new long-term, large-scale frame-event visual object tracking dataset, termed FELT. It contains 1,044 long-term videos that involve 1.9 million RGB frames and event stream pairs, 60 different target objects, and 14 challenging attributes. To build a solid benchmark, we retrain and evaluate 21 baseline trackers on our dataset for future work to compare. In addition, we propose a novel Associative Memory Transformer based RGB-Event long-term visual tracker, termed AMTTrack. It follows a one-stream tracking framework and aggregates the multi-scale RGB/event template and search tokens effectively via the Hopfield retrieval layer. The framework also embodies another aspect of associative memory by maintaining dynamic template representations through an associative memory update scheme, which addresses the appearance variation in long-term tracking. Extensive experiments on FELT, FE108, VisEvent, and COESOT datasets fully validated the effectiveness of our proposed tracker. Both the dataset and source code will be released on \url{https://github.com/Event-AHU/FELT_SOT_Benchmark}
\end{abstract}

\section{Introduction} 
\begin{figure}
\centering
\includegraphics[width=\linewidth]{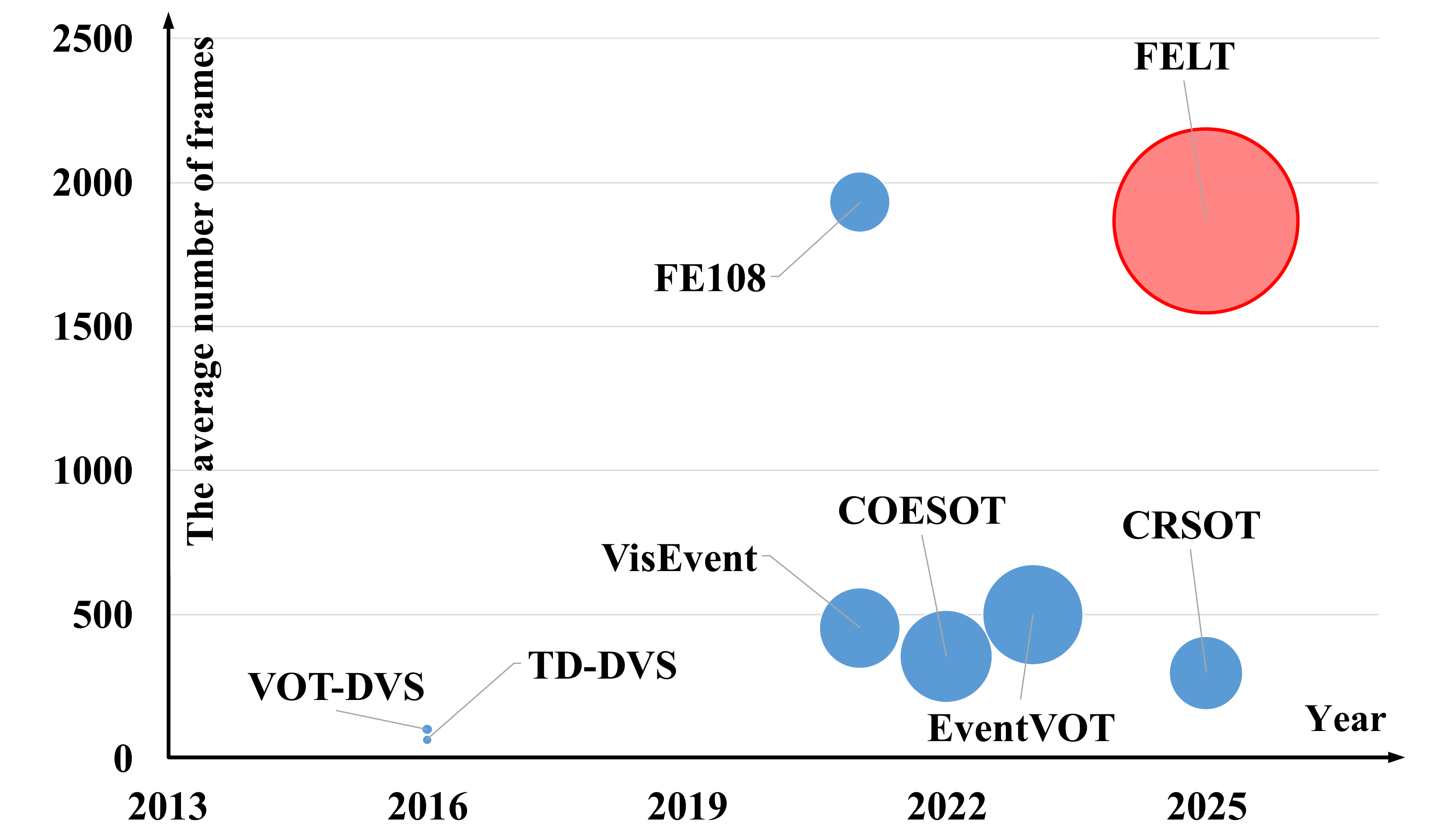}
\caption{Comparison between event-based tracking datasets. The size of the bubbles represents the total number of frames.} 
\label{fig:intro}
\end{figure}

Visual object tracking aims to locate the specified object using an adaptively adjusted bounding box. This task has been widely employed in practical applications, such as automatic piloting, drone aerial photography, and intelligent security monitoring. Usually, the visual trackers are developed and deployed based on RGB cameras, which record the scenario using video frames. As we know, RGB cameras are easily influenced by challenging factors, such as low illumination, fast motion, and heavy occlusion. Thus, RGB cameras based tracking is still a very challenging task even nowadays. 

To address the aforementioned issues, researchers begin to exploit new sensors to replace or assist the RGB cameras for more robust tracking. Among them, bio-inspired event cameras (e.g., DVS346, CeleX-V, Prophesee) have drawn increasing attention in recent years due to their unique imaging principles. Different from RGB cameras, which record all the pixel intensities in the scene in a synchronous way and output video frames at a fixed frame rate (e.g., 30 frames per second), the event cameras emit a spike for each pixel only when the variation of light intensity exceeds a given threshold. The event pixel values (-1 or 1) are independently and asynchronously recorded, with -1 or 1 (corresponding to an OFF or ON event) denoting that the light intensity is decreased or increased. As noted in~\cite{gallego2020event}, event cameras also perform better than the RGB cameras on the \textit{high dynamic range}, \textit{low energy consumption}, and \textit{high temporal resolution}. Thus, it can be used in low-illumination and fast-motion tracking scenarios. Even with the above advantages, relying solely on event cameras still performs poorly in some scenes, such as static or motionless objects, because only the signals of motion can be sensed, and stationary objects cannot be captured. Event cameras also fail to capture the color information or very fine-grained texture information. Fortunately, this is where the RGB cameras excel. Therefore, the usage of both RGB and event cameras for tracking has great landing value and potential applications. 


Based on this idea, there are already several datasets proposed for event-based tracking, including FE108~\cite{zhang2021object}, VisEvent~\cite{wang2023visevent}, COESOT~\cite{tang2022revisiting}, EventVOT~\cite{wang2024event}, and CRSOT~\cite{zhu2024crsot}. 
Specifically, FE108~\cite{zhang2021object} is collected using a gray-scale DVS event camera in the indoor scenarios, and EventVOT~\cite{wang2024event} is a pure event-based tracking dataset. 
VisEvent~\cite{wang2023visevent}, COESOT~\cite{tang2022revisiting}, and CRSOT~\cite{zhu2024crsot} are three large-scale tracking datasets; however, most of their samples are short-term.
Conducting long-term frame-event single object tracking has the following advantages and benefits: 
\textit{1). Long-term RGB-Event tracking is closer to real-world application scenarios:} real surveillance and security systems typically require continuous tracking over extended periods, and military reconnaissance and wildlife monitoring applications need long-term tracking capabilities. 
\textit{2). Greater technical challenges:} it can test the robustness of multi-modal tracking algorithms under long-term occlusion, verify the adaptability to target appearance changes (scale, pose, illumination), and evaluate the handling capability of target disappearance and reappearance. 
In addition, the release of long-term multimodal tracking datasets can also \textit{3). advance research on efficient tracking algorithms}, move beyond frame-by-frame fusion limitations, and reduce the power consumption of multimodal tracking algorithms.

Based on the aforementioned observations and reflections, in this work, we propose the first \textbf{F}rame-\textbf{E}vent \textbf{L}ong-\textbf{T}erm single object tracking benchmark dataset, termed \textbf{FELT}. 
It contains 1,044 videos and 1,949,680 RGB frames and event stream pairs and has become the largest frame-event tracking dataset to date, as shown in Table~\ref{tab:event_datasets} and Fig.~\ref{fig:intro}. 14 challenging attributes (e.g., illumination variation, fast motion, out of view, and occlusion) and 60 classes of target objects are involved in this dataset. 
These videos are split into training and testing subsets, which contain 730 and 314 sequences, respectively. 
In addition, we retrain and evaluate 21 baseline trackers on our dataset to build a high-quality benchmark. We believe the dataset and benchmark results build a good platform for the development of long-term frame-event single object tracking.

In this paper, we also propose a novel associative memory augmented Transformer for the long-term RGB-Event visual object tracking, termed AMTTrack. 
The key insight lies in leveraging associative memory to explore the correlations between features at different levels and modalities, thereby optimizing more expressive feature representations. 
Meanwhile, through associative memory, it aggregates diverse appearance representations of target objects during the tracking process, further alleviating the challenge of modeling object appearance in long-term tracking. 
More specifically, we first adopt the template and search areas corresponding to RGB and Event frames as inputs, and get the token sequences using the projection layer. These tokens are fed into the associative memory augmented Transformer backbone network for simultaneous multi-level feature extraction and interactive learning. 
Then, the tracking head takes the fused search tokens for target localization. 
In addition, we store the target object obtained in the tracking procedure and retrieve dynamic RGB/Event templates to adapt to appearance variation via a Hopfield layer. 
An overview of our proposed AMTTrack can be found in Fig.~\ref{fig:framework}. 

To sum up, the main contributions of this work can be summarized from the following three perspectives:
\begin{itemize}

\item We propose a large-scale and long-term RGB-Event visual object tracking dataset, termed FELT, which contains 1,044 video sequences and benchmark 21 baselines for future works to compare. 

\item We propose a novel associative memory augmented RGB-Event visual object tracking framework, termed AMTTrack. The expressive capability of the Transformer backbone is further enhanced through a carefully designed hierarchical multi-modal feature interaction mechanism. 

\item Extensive experiments on four benchmark tracking datasets (FELT, FE108, VisEvent, COESOT) fully verify the effectiveness of our proposed tracker.
\end{itemize}

\section{Related Works} 
In this section, we review the prominent visual trackers based on Event Camera based Tracking, Event-based datasets, and Modern Hopfield Networks. More related works can be found in the following surveys~\cite{gallego2020event} and paper list~\footnote{\url{https://github.com/wangxiao5791509/Single_Object_Tracking_Paper_List}}. 

\subsection{Event Camera based Tracking} 
As a newly arising research topic in computer vision, Event-based tracking~\cite{zhang2021object} quickly draws more and more attention. 
Wang et al.~\cite{wang2023visevent} propose the Cross-Attention Transformer (CMT) to enable feature interaction between RGB and event data for tracking. 
AFNet~\cite{zhang2023frame} proposed by Zhang et al. introduces a multi-modal alignment and fusion module, which adaptively merges complementary information from high frame rate event streams and RGB frames while attenuating noise.
Wang et al. propose HDETrack~\cite{wang2024event} which utilizes hierarchical knowledge distillation to transfer multi-modal/view knowledge to the single-modal event stream, enabling high-speed and low-latency visual tracking.
Tang et al. propose CEUTrack~\cite{tang2022revisiting} which employs a unified Transformer backbone to model the relationship between RGB frames and event voxels, thereby facilitating effective feature interaction. 
Zhu et al.~\cite{zhu2023cross} introduce a mask modeling strategy aimed at addressing the issue of cross-modal interaction between RGB frames and event streams. 
Fu et al. introduce DANet~\cite{fu2023distractor}, crafting an event-based interference sensing tracker through the integration of Transformer modules and Siamese architecture. 
ViPT~\cite{zhu2023visual} is proposed by Zhu et al. which incorporates learnable modal-relevant prompts while retaining frozen pre-trained models, enhancing the adaptability of the models to diverse multi-modal tracking tasks.
Zhu et al.~\cite{zhu2022learning} directly process event clouds and predict the motion-aware target likelihood. 
Zhang et al. propose a cross-domain attention fusion algorithm in STNet~\cite{zhang2022spiking}. 
SDSTrack~\cite{hou2024sdstrack} proposed by Hou et al. uses complementary masked patch distillation to boost extreme conditions tracking.
Sun et al. introduce MamTrack~\cite{sun2025exploring}, which leverages Mamba for long-range historical modeling.
Different from these works, we propose a novel long-term RGB-Event tracker using the associative memory Transformer.

\begin{table*}
\centering
\footnotesize
\caption{Comparison of Event camera-based visual tracking datasets.} 
\label{tab:event_datasets}
\setlength{\tabcolsep}{1mm} 
\begin{tabular}{l|cccccccccccc}
\toprule
\textbf{Datasets} & \textbf{Year} &\textbf{Videos}  & \textbf{Frames}  &\textbf{Class}   &\textbf{Attributes} &\textbf{Resolution} &\textbf{Aim} &\textbf{Absent} &\textbf{Frame} &\textbf{Reality} &\textbf{LT} &\textbf{Public}\\
\hline 
\textbf{VOT-DVS}~\cite{hu2016dvs} & 2016 & 60 & - & - & - & 240 × 180 & Eval & \ding{55} & \ding{55} & \ding{55} & \ding{55} & \ding{51} \\
\textbf{TD-DVS}~\cite{hu2016dvs} & 2016 & 77 & - & - & - & 240 × 180 & Eval & \ding{55} & \ding{55}  & \ding{55} & \ding{55} & \ding{51} \\
\textbf{Ulster}~\cite{liu2016combined} & 2016 & 1 & 9,000 & - & - & 240 × 180  & Eval & \ding{55} & \ding{55} & \ding{51} & \ding{51} & \ding{55} \\
\textbf{EED}~\cite{mitrokhin2018event} & 2018 & 7 & 234 & - & - & 240 × 180  & Eval & \ding{55} & \ding{55} & \ding{51} & \ding{55} & \ding{51} \\
\hline
\textbf{FE108}~\cite{zhang2021object} & 2021 & 108 & 208,672 & 21 & 4 & 346 × 260 & Train \& Eval & \ding{55} & \ding{55} & \ding{51} & \ding{51} & \ding{51} \\
\textbf{VisEvent}~\cite{wang2023visevent} & 2021 & 820 & 371,127 & - & \textbf{17} & 346 × 260 & Train \& Eval & \ding{51} & \ding{51} & \ding{51} & \ding{55} & \ding{51} \\
\textbf{COESOT}~\cite{tang2022revisiting} & 2022 & \textbf{1,354} & 478,721 & \textbf{90} & \textbf{17} & 346 × 260 & Train \& Eval & \ding{51} &\ding{51} & \ding{51} & \ding{55} & \ding{51}  \\
\textbf{EventVOT}~\cite{wang2024event} & 2023 & 1,141 & 569,359 & 19 & 14 & 1280 × 720 & Train \& Eval & \ding{51} & \ding{55} & \ding{51} & \ding{55} & \ding{51}  \\
\textbf{CRSOT}~\cite{zhu2024crsot} & 2025 & 1,030 & 304,974 & - & \textbf{17} & \textbf{1280 × 800} & Train \& Eval & \ding{51} & \ding{51} & \ding{51} & \ding{55} &
\ding{51} \\
\hline
\textbf{FELT (Ours)} & 2025 & 1,044 & \textbf{1,949,680} & 60 & 14 &  346 × 260 & Train \& Eval & \ding{51} & \ding{51} & \ding{51} & \ding{51} & \ding{51} \\
\bottomrule
\end{tabular} 
\end{table*}

\subsection{Benchmark Datasets for Event-based Tracking} \label{eventdatasets} 
Event stream-based visual tracking has garnered increasing attention in recent years, highlighting the growing importance of event-based benchmark datasets. To uphold dataset quality, manual labeling of target boxes is often a requisite for event-based benchmark datasets. Existing widely used event-based single object tracking datasets include FE108~\cite{zhang2021object}, VisEvent~\cite{wang2023visevent}, COESOT~\cite{tang2022revisiting}, EventVOT~\cite{wang2024event} and CRSOT~\cite{zhu2024crsot}. 

\noindent \textbf{FE108}~\cite{zhang2021object} comprises 108 videos captured under consistent lighting conditions, totaling 208,672 frames for the entire dataset, averaging 1,932 frames per video. The dataset encompasses 21 distinct categories of target objects and introduces four challenging factors. 

\noindent \textbf{VisEvent}~\cite{wang2023visevent} encompasses 820 video sequences, totaling 371,127 frames, with an average length of 453 frames per video. Six fundamental guidelines were adhered to during the dataset collection process to ensure high-quality, diverse, and large-scale datasets. The dataset incorporates 17 challenging factors, encompassing RGB mode challenges such as fast motion, low illumination, and exposure, as well as event mode challenges like the absence of target movement, background confusion, etc.

\noindent \textbf{COESOT}~\cite{tang2022revisiting} is a large-scale benchmark dataset for color-event tracking. It comprises 1,354 video sequences encompassing 90 target object categories. The dataset spans a total of 478,721 frames, averaging 354 frames per video. The diverse video scenes include indoor, street, zoo, and various other environments. Noteworthy challenges within the dataset involve factors such as the rotation of target objects, complete occlusion, and movement of background objects. 

\noindent \textbf{EventVOT}~\cite{wang2024event} stands as the pioneering high-resolution, large-scale, single-modal event dataset. Encompassing 1,141 video sequences, the dataset comprises a total of 569,359 frames, averaging 499 frames per video. Capturing a diverse range of scenes, including indoor, street, and stadium, the videos span both day and nighttime periods. The dataset prominently features 14 challenging factors, such as speed changes and variations in lighting conditions. Notably, all data undergo annotation by a professional labeling company to ensure the dataset's overall high quality. 

\noindent \textbf{CRSOT}~\cite{zhu2024crsot} 
is the first unaligned frame-event dataset. It employs a hybrid camera system to capture high-resolution data (1440×1080 RGB, 1280×800 CeleX-V), with the NIR camera assisting low-light annotation.
With 1,030 RGB-Event pairs and 304,974 frames, it captures DVS features, and defines 17 attributes for tracker performance evaluation. 
Key challenges include background clutter, background object motion, low illumination, and partial occlusion.

Compared with these datasets, our proposed FELT is the \textbf{first} \textbf{long}-\textbf{term} frame-event tracking dataset, which is the \textbf{largest} one, as shown in Table~\ref{tab:event_datasets}. 

\subsection{Modern Hopfield Networks}
Hopfield network~\cite{hopfield1982neural} proposed by John Hopfield in 1982 is well known for its associative memories, and it is treated as one of the earliest artificial neural models. This neural network can associate an input with its most similar pattern, i.e., retrieve a stored memory based on partial or incomplete input. 
Different from the classical binary Hopfield networks, recent works~\cite{krotov2016dense, demircigil2017model} demonstrate that by introducing polynomial or exponential terms into the energy function, the storage capacity can be increased significantly. Modern Hopfield networks~\cite{ramsauer2020hopfield} \footnote{\url{https://ml-jku.github.io/hopfield-layers/}} proposed by Ramsauer et al. achieve continuous associative memory. After that, many tasks are boosted with the help of modern Hopfield networks, such as remote sensing image generation~\cite{xu2023txt2img}, reaction template prediction~\cite{seidl2022improving}, immune repertoire classification~\cite{widrich2020modern}, nuclear fusion~\cite{ma2024exploiting}, and X-ray report generation~\cite{wang2025activating}.
Inspired by the strong associative memory of modern Hopfield networks, we introduce associative memory to model correlations in long-term RGB-Event tracking.

\section{FELT Single Object Tracking Dataset}

\begin{table}
\center
\small  
\caption{Description of 14 attributes in our FELT dataset.} 
\label{tab:felt_attributes}
\setlength{\tabcolsep}{1mm} 
\begin{tabular}{l|l}
\toprule
\textbf{Attributes} & \textbf{Description}  \\ 
\hline
\textbf{01. ST} & The scale of the target object is limited \\	
\textbf{02. OE} & The light intensity is very high \\	
\textbf{03. IV} & The light intensity changes during tracking \\
\textbf{04. NMO} & The target object is stationary \\
\textbf{05. ARC} & The ratio of the bounding box aspect ratio varied \\	
\textbf{06. BI} & The target object is influenced by the background \\ 
\textbf{07. SV} & The width and height of the target object changed \\ 
\textbf{08. POC} & Part of the target object is occluded  \\
\textbf{09. OV} & The target object moves out of the view \\
\textbf{10. LI} & The videos are recorded in the dark environment  \\
\textbf{11. FOC} & The target object is fully occluded by other objects \\
\textbf{12. DEF} & The shape of the target object changed \\ 
\textbf{13. VT} & The views of the target object vary during tracking \\ 
\textbf{14. FM} & The target object moves quickly \\ 
\bottomrule
\end{tabular}
\end{table}		


\begin{figure*}
\centering
\includegraphics[width=\linewidth]{figures/attributes_and_bbox.jpg}
\caption{(a). The number of videos corresponding to specific challenging factors; (b). Distribution of bounding box.}
\label{fig:felt_attributes_and_bbox}
\end{figure*}

\subsection{Design Principle} 
We follow the following principles when collecting the FELT SOT dataset: 
\textit{1). Long-term}: Each video contains at least 1000 video frames and event streams. 
\textit{2). Large-scale}: To bridge the gap between data-hungry deep neural networks, we aim to build the largest frame-event tracking dataset. 
\textit{3). Various Challenges}: The collected videos need to reflect the key challenges in frame-event tracking. 
\textit{4). Dual-Modality}: To meet the demand for event-only and frame-event fusion-based tracking tasks, we need to build a dataset with both RGB and Event data. 
\textit{5). Incomplete Information}: Considering the particularity of multi-modal tracking, the dataset needs to take into account the incomplete information of each modality. 
For example, the event data is spatially sparse and the RGB data is weakened.

\subsection{Dataset Collection and Annotation} 
The DVS346 event camera, which outputs the RGB frames (346×260) and event streams simultaneously, is adopted for FELT dataset collection. 
The two modalities are aligned well from a hardware perspective on both spatial and temporal views. 
When recording our videos, we follow the design principles and challenging attributes defined in subsection 3.1 and 3.3.
For the bounding box annotation, we employ a professional data company to annotate and check for several rounds to ensure quality.

\begin{figure*}
\centering
\includegraphics[width=1\linewidth]{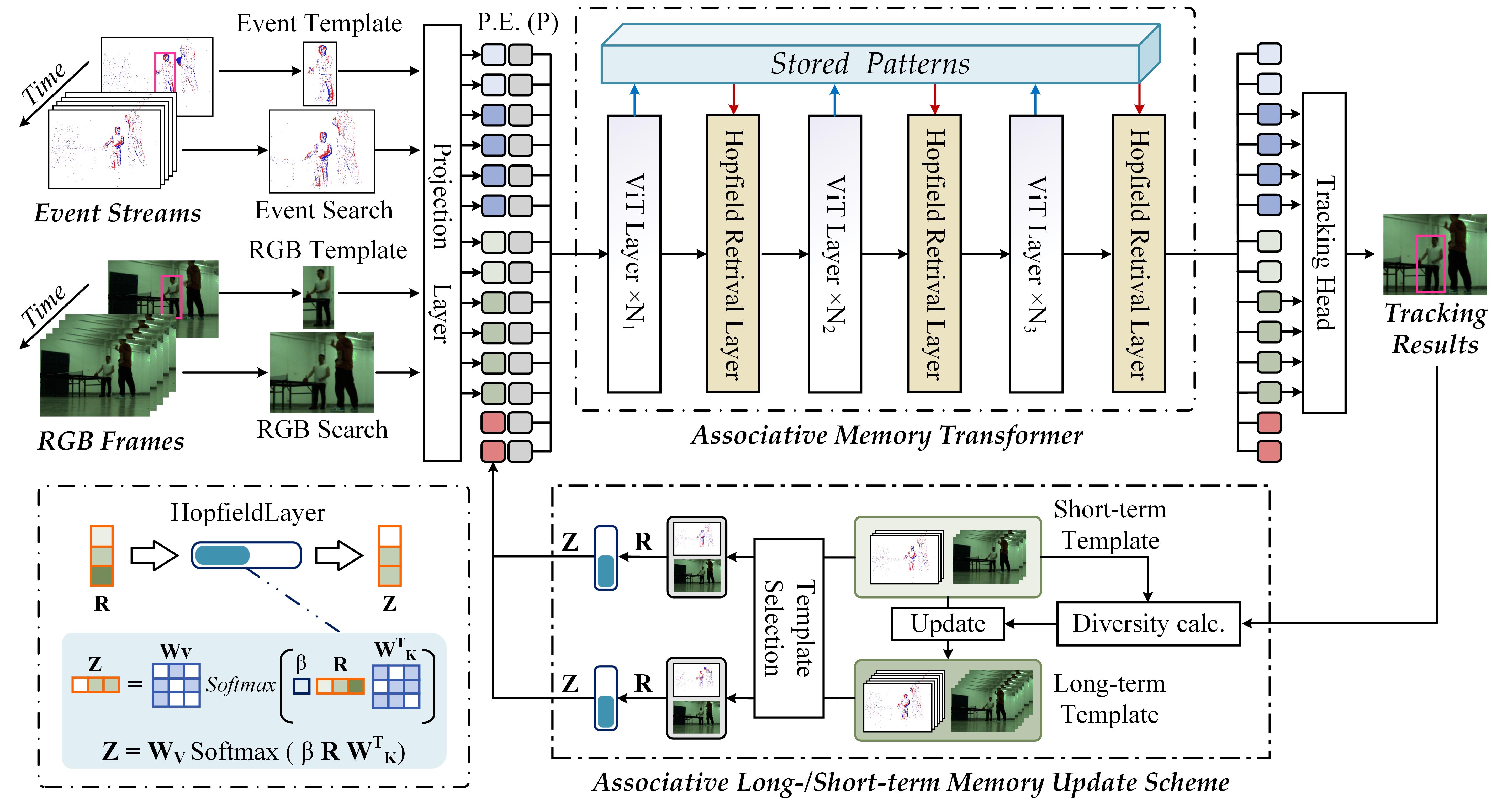}
\caption{An overview of our proposed associative memory augmented RGB-Event visual tracking framework. Our tracker (termed AMTTrack) is composed of three parts: Input Embedding Layer, Associative Memory Transformer based on Retrieval Augmentation, and Associative Long-/Short-term Memory Update Scheme.} 
\label{fig:framework}
\end{figure*}

\subsection{Attributes} 
When collecting the FELT SOT dataset, we consider 14 attributes, as shown in Table~\ref{tab:felt_attributes}, for the long-term frame-event tracking task. 
Specifically, the ST (Small Target) and illumination related attributes (OE (Over Exposure), IV (Illumination Variation), LI (Low Illumination)) are all considered. 
We also define the motion-based attributes, like the NMO (No MOtion) and FM (Fast Motion). The OV (Out of View) is considered for the long-term tracking task. 
We believe these defined challenging attributes will be a better platform for the evaluation of future trackers.
The number of videos corresponding to these attributes is visualized in Fig.~\ref{fig:felt_attributes_and_bbox} (a). We can find that BI, OV, and VT are top-3 major challenges which demonstrate that our dataset is relatively challenging.

\subsection{Statistical Analysis} 
As shown in Table~\ref{tab:event_datasets}, our dataset contains 1,044 videos, 1,949,680 RGB frames, which cover 60 categories of target objects in both indoor and outdoor scenarios. 
We split these videos into training and testing subsets, which contain 730 and 314 videos, respectively. 
In addition, the distribution of the center points of the annotated bounding boxes is visualized in Fig.~\ref{fig:felt_attributes_and_bbox} (b).



\section{Methodology} 

\subsection{Overview} 
As shown in Fig.~\ref{fig:framework}, we first use the templates and search areas corresponding to RGB and Event frames as inputs, and after encoding through the projection layer and adding position embedding, the input is converted into a token sequence. Then we input it into the unified ViT~\cite{dosovitskiy2020image} backbone for relational modeling. 
We embed Hopfield retrieval layers to further explore multi-level feature extraction and enhance the representation ability of each modality. 
Finally, the fused search area features are output using the tracking head for object positioning, and the tracking results are optionally stored in memory into the short-term and long-term template memory, and the templates are dynamically updated for subsequent frames to adapt to the object changes during the long-term tracking process. 
More details will be introduced separately in the following sections.

\begin{figure*}
\centering
\includegraphics[width=1\linewidth]{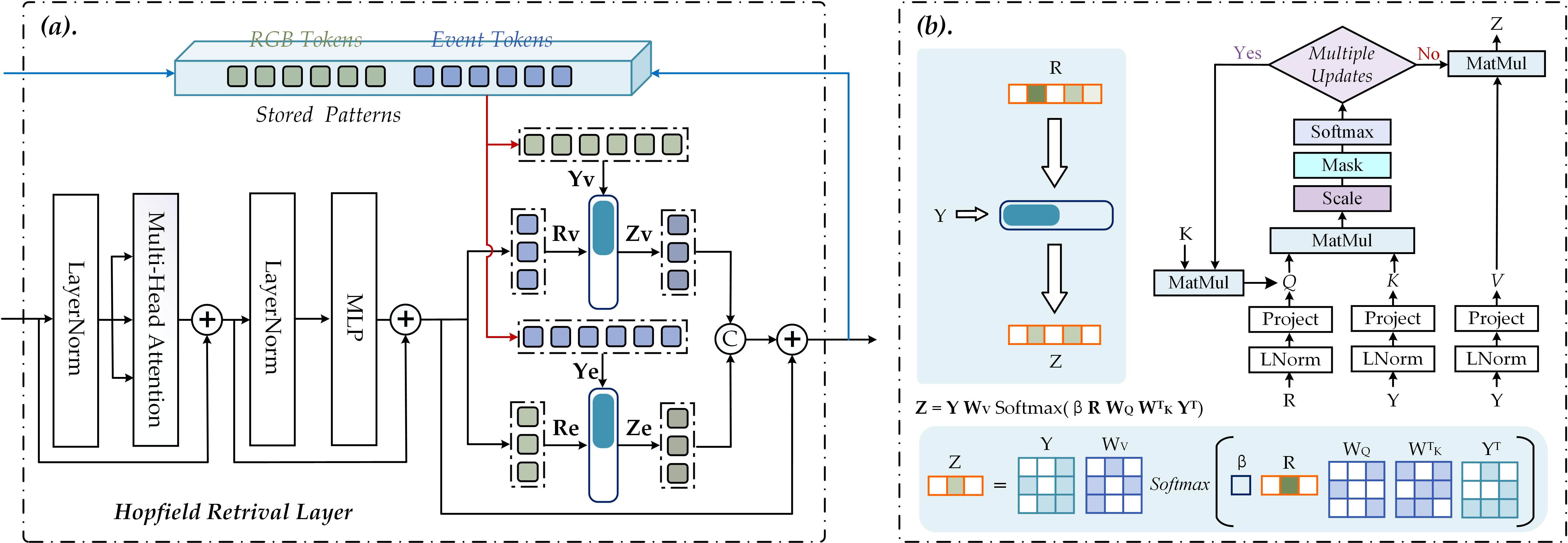}
\caption{Detailed network architecture of (a). Hopfield Retrieval Layer; (b). Modern Hopfield Network. }
\label{fig:hopfieldNet}
\end{figure*}

\subsection{Input Representation} 
The input of our AMTTrack framework consists of template and search area images from RGB and Event, where the sizes of the template and search area are $128\times128$ and $256\times256$, respectively.
We denote the RGB frames as $\mathcal{I}_{i} \in \mathbb{R}^{C\times H\times W}$ and the Event frames as $\mathcal{E}_{i} \in \mathbb{R}^{C\times H\times W}$, where $i$ is the index of RGB-Event image pairs. 
The event frames $\mathcal{E}_i$ are composed of event streams $\{e_1, e_2,\cdots,e_M\}$ stacked at a fixed interval, where $e_j=[x,y,t,p]$ represents each event point.
Then, we first perform patch embedding and add position encoding on template frames and search frames to obtain template and search token sequences of RGB and Event, which are RGB template token embeddings $\mathcal{Z}_{v} \in \mathbb{R} ^{N_z \times D}$, RGB search token embeddings $\mathcal{X}_{v} \in \mathbb{R} ^{N_x \times D}$, Event template token embeddings $\mathcal{Z}_{e} \in \mathbb{R} ^{N_z \times D}$, and Event search token embeddings $\mathcal{X}_{e} \in \mathbb{R} ^{N_x \times D}$, respectively.
Here, $N_z = \frac{H_z\times{W_z}}{P^2}$, $N_x = \frac{H_x\times{W_x}}{P^2}$ , and $P=16$.

\subsection{Associative Memory Augmented Hopfield} 
In this subsection, we elaborate on the design and implementation of the Associative Memory Augmented Hopfield module within our RGB-Event visual tracking framework. 
The module employs the Hopfield network for feature retrieval, thereby enhancing the model's hierarchical feature extraction and representation ability.


\textbf{Modern Hopfield Networks.}
According to \cite{ramsauer2020hopfield}, modern Hopfield networks converge to a steady state by minimizing an energy function corresponding to stored memory patterns. By adjusting the weight matrix, the network can store multiple patterns, with each pattern corresponding to a steady state. Its energy function is expressed as:
\begin{equation}
    E = -\frac{1}{\beta} \log \sum_{i = 1}^{N} \exp (\beta \bm{y}_i^{T} \bm{r} ) + \frac{1}{2}\left \| \bm{r} \right \|^2 + C,
\end{equation}
where $\beta$ denotes a temperature parameter, $\bm{y}_i$ represents the $i$-th stored pattern, $\bm{r}$ denotes the input state pattern, and $N$ stands for the number of stored patterns. $C$ represents a constant quadratic term.
Inspired by \cite{ramsauer2020hopfield}, modern Hopfield networks exhibit a strong connection with attention mechanisms, and the update rule of the new energy function is the self-attention of transformer networks. It can be formally expressed as:
\begin{equation}
    \bm{r}^{(t+1)} = \bm{Y} \cdot \operatorname{softmax}( \beta \bm{Y} \bm{r}^{(t)} ),
\end{equation}
where $\bm{Y}=(\bm{y}_1, \dots, \bm{y}_N)^T$ denotes the memory bank, and $\bm{r}^{(t)}$ represents the input state pattern at time $t$.
Further, we map each pattern to the associative space through projection matrices:
\begin{equation}
\label{eq:hopfield}
    \bm{Z} = \bm{Y} \bm{W}_V \cdot \operatorname{softmax} \left( \beta \bm{R} \bm{W}_Q \bm{W}^T_K \bm{Y}^T \right), 
\end{equation}
where $\bm{Z}$ denotes the result patterns, $\bm{R}$ represents the state patterns, and $\bm{Y}$ means the stored patterns. The projection matrices $\bm{W}_Q$, $\bm{W}_K$, and $\bm{W}_V$ transform these patterns. 
The temperature $\beta$ controls the sharpness of the softmax distribution. A high value of $\beta$ indicates that each pattern is easier to retrieve individually. A low value of $\beta$ indicates a higher likelihood of metastability, and the retrieved state will be a superposition of multiple stored patterns. 

\textbf{Hopfield Embedding in Transformer Layers.}
Our network consists of a 12-layer Transformer backbone. 
As shown in Fig.~\ref{fig:hopfieldNet} (a), we embed Hopfield layers into specific layers of the Transformer. These Hopfield layers capture different semantic levels during feature extraction and interaction.
Let $\mathcal{L} = \lbrace l_1, l_2, \dots, l_k \rbrace$ represent the set of Hopfield layers in a Transformer architecture, where each $l_i$ denotes the positional index of the $i$-th Hopfield layer within the 12-layer Transformer.
For each Hopfield layer $l_i\in\mathcal{L}$, we denote the features output from a set of preceding transformer layers $\mathcal{P}_i= \lbrace p_{i1}, p_{i2}, \cdots, p_{im} \rbrace$ as the memory bank of Hopfield, i.e. the stored patterns $\bm{Y}_i$, where $p_{ij}$ represents the index of the $j$-th preceding transformer layer corresponding to $l_i$.
The output of the current transformer layer denoted as $\bm{R}_i$ will be utilized for retrieval of the stored patterns $\bm{Y}_i$.

\textbf{Cross-Modal Retrieval.}
With dual-modal token sequences, we introduce a cross-modal retrieval mechanism, as shown in Fig.~\ref{fig:hopfieldNet} (a), which fully explores the complementary information between different modalities.
In our cross-modal retrieval process, we leverage the Hopfield layer to enhance the feature representation.
Specifically, for each Hopfield layer $l_i\in\mathcal{L}$, when $\bm{R}_{v}^i$ is the RGB search feature of the $l_i$-th layer, we employ Event features from the layers in $\mathcal{P}_i$ as $\bm{Y}_{e}^i$ for retrieval. 
Conversely, when $\bm{R}_{e}^i$ is the Event search feature from the $l_i$-th layer, we employ the RGB features from the layers in $\mathcal{P}_i$ as $\bm{Y}_{v}^i$ for retrieval.
Mathematically, the cross-modal retrieval process can be formulated as follows:
\begin{equation}
    \begin{cases}
        \bm{Z}_{v}^i = Hopfield(\bm{R}_{v}^i, \bm{Y}_{e}^i) + \bm{R}_{v}^i, \\
        \bm{Z}_{e}^i = Hopfield(\bm{R}_{e}^i, \bm{Y}_{v}^i) + \bm{R}_{e}^i,
    \end{cases}
\end{equation}
where $\bm{Z}_{v}^i$ and $\bm{Z}_{e}^i$ are the output features after cross-modal retrieval, and $Hopfield(\cdot,\cdot)$ represents the Hopfield layer. 

Specifically, through the cross-modal retrieval mechanism described above, it enables our model to effectively integrate multi-modal information, enhancing tracking accuracy and robustness, especially in complex scenarios with insufficient single-modality data. 
Moreover, the retrieved patterns from multi-layer dual-modal features further integrate local details and global context from different Transformer layers.

\subsection{Associative Template Update}
Long-term visual tracking may cause drastic object appearance changes, making it hard to localize the target accurately with only the initial static template. 
We introduce an associative memory-based dynamic template updating strategy, which maintains a short-term template memory and a long-term template memory, and design a dynamic template updating mechanism based on appearance diversity matching for dynamically updating the template memory during short- and long-term visual tracking.

\textbf{Long-/Short-term Template Memory.}
Inspired by THOR~\cite{sauer2019tracking} and MambaEVT~\cite{wang2025mambaevt}, we use short-term (ST) and long-term (LT) memories to store templates as cropped images from previous predictions.
In detail, the ST maintains a dynamic template bank, updated when two conditions are met: (1) the predefined short-term update interval arrives, and (2) the response map's maximum confidence score surpasses a specified threshold.
Based on these rules, we introduce an appearance-diversity-based long-term template update mechanism. It enriches the long-term template memory with diverse features to adapt to object appearance variability.
Moreover, we initialize the long-term template memory at certain intervals to prevent the accumulation of errors in the long-term process from causing the characteristics of the long-term template memory to differ too much from the current frame object.

\textbf{Associative Hopfield Augmented Template.}
In complex scenarios, even optimal template updates can add low-quality templates to memory, ultimately leading to tracking performance degradation. 
To address this, we introduce association memory-augmented dynamic templates to optimize their representation.
As shown in Fig.~\ref{fig:framework}, the Hopfield layer uses static stored and state patterns. It does not depend on the network input; instead, it is determined solely by the bias weights (i.e., no additional stored patterns $\bm{Y}$ are required).
The Hopfield layer can be expressed as follows:
\begin{equation}
    \bm{Z} = \bm{W}_V \cdot \operatorname{softmax}(\beta\bm{R}\bm{W}^T_K).
\end{equation}
Specifically, we use the Hopfield layer to store learnable prototypes. During training, dynamic templates serve as stored patterns. For inference, we select dynamic templates from memory as state patterns $\bm{R}$. The Hopfield layer's lookup mechanism outputs updated dynamic templates $\bm{Z}$, enhancing their representation.

\textbf{Update Process.}
The specific dynamic template update process is shown in Algorithm~\ref{alg:dynamic_template_update}. The input comprises the initial static template $\mathbf{z}_s$, the dynamic template $\mathbf{z}_d$, and the response $\mathbf{score}$. The parameters comprise the short-term update interval $I_{su}$, the long-term update interval $I_{lu}$, the similarity lower bound $\theta$, and the response threshold $\tau$.

\begin{algorithm}[tb]
\caption{Algorithms for Dynamic Template Update}
\label{alg:dynamic_template_update}
\textbf{Input}: $\mathbf{z}_{s}$, $\mathbf{z}_{d}$, $\mathbf{score}$ \\
\textbf{Parameter}: $I_{su}$, $I_{lu}$, $\theta$, $\tau$ \\
\textbf{Output}: Dynamic templates $\mathbf{z}$ 
\begin{algorithmic}[1]
    \STATE \textbf{Init}: Use $\mathbf{z}_{s}$ to init short/long-term memories; $n \gets 0$
    \FOR{each frame}
        \STATE $n \gets n + 1$
        \IF{$\mathbf{score} > \tau$ and $n \bmod I_{su} = 0$}
            \STATE $\text{div\_scale} \gets \text{Update ST memories}(\mathbf{z}_{d})$
            \IF{$n \bmod I_{lu} = 0$}
                \STATE Update LT memories $(\mathbf{z}_{d}, \text{div\_scale}, \theta)$
            \ENDIF
        \ENDIF
        \STATE $\mathbf{z}_{d}^{LT}, \mathbf{z}_{d}^{ST} \gets \text{Resample dynamic templates}$
        \STATE $\mathbf{z}_{d}^{LT}, \mathbf{z}_{d}^{ST} \gets \text{HopfieldLayer}(\mathbf{z}_{d}^{LT}, \mathbf{z}_{d}^{ST})$
        \STATE $\mathbf{z} \gets \text{Concat}([\mathbf{z}_{d}^{LT}, \mathbf{z}_{d}^{ST}])$
        \STATE \textbf{Return} $\mathbf{z}$
    \ENDFOR
\end{algorithmic}
\end{algorithm}

\subsection{Tracking Head and Loss Function}
After obtaining dual-modal search region fusion features, we input it into the tracking head for target object localization. The tracking head adopts the same configuration as OSTrack~\cite{ye2022joint}, which is implemented as a Fully Convolutional Network (FCN), and outputs the ultimate predicted bounding box of the object.

In the training phase, the overall loss function of our framework includes L1 loss, GIOU loss~\cite{rezatofighi2019generalized}, and weighted focal loss~\cite{law2018cornernet}.
The loss function can be formulated as:
\begin{equation}
    \label{lossFunction} 
    \mathcal{L}=\lambda_{L_1}L_1 + \lambda_{iou}L_{iou} + L_{focal},
\end{equation}
where $\lambda_{L_1}=5.0$ and $\lambda_{iou}=2.0$ are the weight coefficients in our experiment.

\begin{table*}[h]
\centering
\small
\caption{Experimental results (SR/PR) on FE108 dataset.} 
\label{tab:fe108_results}
\begin{tabular}{cccccccc}
\toprule
\textbf{SiamRPN}~\cite{li2018high} & \textbf{SiamBAN}~\cite{chen2020siamese} & \textbf{SiamFC++}~\cite{xu2020siamfc++} & \textbf{KYS}~\cite{bhat2020know} & \textbf{CLNet}~\cite{dong2020clnet} & \textbf{CMT-MDNet}~\cite{wang2023visevent} & \textbf{ATOM}~\cite{danelljan2019atom} \\
21.8/33.5 & 22.5/37.4 & 23.8/39.1 & 26.6/41.0 & 34.4/55.5 & 35.1/57.8 & 46.5/71.3 \\ 
\midrule
\textbf{DiMP}~\cite{bhat2019learning} & \textbf{PrDiMP}~\cite{danelljan2020probabilistic} & \textbf{CMT-ATOM}~\cite{wang2023visevent} & \textbf{CEUTrack}~\cite{tang2022revisiting} & \textbf{ViPT}~\cite{zhu2023visual} & \textbf{MamTrack}~\cite{sun2025exploring} & \textbf{Ours} \\
52.6/79.1 & 53.0/80.5 & 54.3/79.4 & 55.58/84.46 & 65.8/93.8 & \textbf{66.4}/94.2 & 65.6/\textbf{95.9} \\
\bottomrule
\end{tabular}
\end{table*}

\section{Experiments} 
\subsection{Datasets and Evaluation Metric}  
To evaluate the effectiveness of our proposed framework, extensive experiments are conducted on four benchmark datasets: \textbf{FE108}, \textbf{FELT}, \textbf{COESOT}, and \textbf{VisEvent}. 
The datasets mentioned above contain different RGB-Event challenge scenarios and cover various complex factors that arise in the short-term and long-term tracking process.

For the evaluation metrics, we employ three widely used metrics to evaluate the results of tracking: \textbf{Success Rate (SR)}, \textbf{Precision (PR)}, and \textbf{Normalized Precision (NPR)}.

\subsection{Implementation Details}  
In the training phase, we follow the settings of OSTrack~\cite{ye2022joint} and use the AdamW optimizer. The learning rate is set to 4e-4, the weight decay is set to 1e-4, and the batch size is set to 12.
We sample templates and search images in a 100-frame window, with 20,000 samples per epoch.
In the testing phase, ST and LT are initialized with an initial static template at the beginning and dynamically updated with tracking results in subsequent phases. The default capacity for ST and LT is 5 and 10, respectively.

\subsection{Comparison on Public Benchmarks}

\noindent $\bullet$ \textbf{Results on FE108 Dataset.} 

FE108~\cite{zhang2021object} includes 108 long-term video sequences, with an average of 1,932 frames per video, making it suitable for evaluating long-term tracking scenarios.
As shown in Table~\ref{tab:fe108_results}, we report the results of our trackers and other existing SOTA methods on FE108.
This result proves that our tracker achieves competitive performance. With SR and PR at 65.6 and 95.9, respectively, the PR surpasses all comparable trackers.

\noindent $\bullet$ \textbf{Results on FELT Dataset.}
\begin{table}[!tp]
\centering
\small
\caption{Experimental results on FELT dataset.} 
\label{tab:felt_results}
\setlength{\tabcolsep}{4pt}
\begin{tabular}{l|c|ccc|c}
\toprule
\textbf{Trackers} & \textbf{Source} &\textbf{SR}  &\textbf{PR} &\textbf{NPR} &\textbf{FPS} \\
\hline 
\textbf{01. Stark~\cite{yan2021learning}} & ECCV22 & 52.7 & 67.9 & 62.8 & 42 \\ 
\textbf{02. OSTrack~\cite{ye2022joint}} & ECCV22 & 52.3 & 65.9 & 63.3 & 75 \\ 
\textbf{03. MixFormer~\cite{cui2022mixformer}} & CVPR22 & 53.0 & 67.5 & 63.8 & 36 \\ 
\textbf{04. AiATrack~\cite{gao2022aiatrack}} & ECCV22 & 52.2 & 66.7 & 62.8 & 31 \\ 
\textbf{05. SimTrack~\cite{chen2022backbone}} & ECCV22 & 49.7 & 63.6 & 59.8 & 82 \\ 
\textbf{06. GRM~\cite{gao2023generalized}} & CVPR23  & 52.1 & 65.6 & 62.9 & 38 \\ 
\textbf{07. ROMTrack~\cite{cai2023robust}} & ICCV23  & 51.8 & 65.8 & 62.7 & 64 \\ 
\textbf{08. ViPT~\cite{zhu2023visual}} & CVPR23 & 52.8 & 65.3 & 63.1 & 29  \\ 
\textbf{09. SeqTrack~\cite{chen2023seqtrack}} & CVPR23  & 52.7 & 66.9 & 63.4 & 31 \\ 
\textbf{10. ARTrackv2~\cite{bai2024artrackv2}} & CVPR24 & 52.3 & 65.2 & 62.8 & 42 \\ 
\textbf{11. HIPTrack~\cite{cai2024hiptrack}} & CVPR24 & 51.6 & 65.6 & 62.2 & 39 \\ 
\textbf{12. ODTrack~\cite{zheng2024odtrack}} & AAAI24 & 52.2 & 66.0 & 63.5 & 57 \\ 
\textbf{13. EVPTrack~\cite{shi2024explicit}} & AAAI24 & 53.8 & 68.7 & 64.8 & 45 \\ 
\textbf{14. AQATrack~\cite{xie2024autoregressive}} & CVPR24 & 54.0 & \textbf{69.1} & 64.7 & 41 \\ 
\textbf{15. SDSTrack~\cite{hou2024sdstrack}} & CVPR24 & 53.7 & 66.4 & 64.1 & 28 \\ 
\textbf{16. UnTrack~\cite{wu2024single}} & CVPR24 & 53.6 & 66.0 & 63.9 & 12 \\ 
\textbf{17. FERMT~\cite{zheng2024exploring}} & ECCV24 & 51.8 & 66.1 & 62.9 & 70 \\ 
\textbf{18. LMTrack~\cite{xu2025less}} & AAAI25 & 50.9 & 63.9 & 61.8 & 72 \\ 
\textbf{19. AsymTrack~\cite{zhu2025two}} & AAAI25 & 51.9 & 66.7 & 62.0 & 104 \\ 
\textbf{20. ORTrack~\cite{wu2025learning}} & CVPR25 & 48.4 & 61.7 & 59.2 & 90 \\ 
\textbf{21. UNTrack~\cite{qin2025must}} & CVPR25 & 50.0 & 63.9 & 61.6 & 23 \\ 
\hline
\textbf{22. Ours} & - & \textbf{54.8} & 67.9 & \textbf{65.7} & 23 \\ 
\bottomrule
\end{tabular} 
\end{table}
FELT dataset contains 1044 long-term video sequences with an average of about 1867 frames per video.
We retrain multiple SOTA trackers and report their performance on the FELT dataset. 
As shown in Table~\ref{tab:felt_results}, compared to our baseline OSTrack, our tracker with the associative memory-based Hopfield layer and template update improves the performance on the FELT dataset, achieving SR 54.8, PR 67.9, and NPR 65.7. 
Notably, compared with other trackers, our method achieves the highest results in both SR and NPR.
However, the PR metric is lower than that of AQATrack~\cite{xie2024autoregressive} and EVPTrack~\cite{shi2024explicit}, which leverage temporal information to capture historical features.

\noindent $\bullet$ \textbf{Results on COESOT Dataset.}
\begin{table}
\centering
\small
\caption{Experimental results on COESOT dataset.} 
\label{tab:coesot_results}
\begin{tabular}{l|c|cc}
\toprule
\textbf{Trackers} & \textbf{Source} &\textbf{SR}  &\textbf{PR} \\
\hline
\textbf{01. Stark~\cite{yan2021learning}} & ICCV21 & 56.0 & 67.7 \\
\textbf{02. KeepTrack~\cite{mayer2021learning}} & ICCV21 & 59.6 & 70.9 \\
\textbf{03. TrDiMP~\cite{wang2021transformer}} & CVPR21 & 60.1 & 72.2 \\
\textbf{04. TransT~\cite{chen2021transformer}} & ECCV22 & 60.5 & 72.4 \\
\textbf{05. OSTrack~\cite{ye2022joint}} & ECCV22 & 59.0 & 70.7 \\
\textbf{06. AiATrack~\cite{gao2022aiatrack}} & ECCV22 & 59.0 & 72.4 \\
\textbf{07. MixFormer~\cite{cui2022mixformer}} & CVPR22 & 55.7 & 66.3 \\
\textbf{08. ToMP101~\cite{mayer2022transforming}} & CVPR22 & 59.9 & 71.6 \\
\textbf{09. MDNet~\cite{wang2023visevent}} & TCYB23 & 53.3 & 66.5 \\
\textbf{10. ViPT~\cite{zhu2023visual}} & CVPR23 & 68.3 & 81.0 \\
\textbf{11. SDSTrack~\cite{hou2024sdstrack}} & CVPR24 & 66.7 & 79.7 \\
\textbf{12. UnTrack~\cite{wu2024single}} & CVPR24 & 67.9 & 80.9 \\
\textbf{13. CMDTrack~\cite{zhang2025cross}} & TPAMI25 & 65.7 & 74.8 \\
\textbf{14. LMTrack~\cite{xu2025less}} & AAAI25 & 58.4 & 71.1 \\
\hline
\textbf{15. Ours} & - & \textbf{68.8} & \textbf{82.9} \\
\bottomrule
\end{tabular} 
\end{table}
COESOT~\cite{tang2022revisiting} is a large-scale benchmark dataset for color-event tracking, which includes 1,354 video sequences and spans a total of 478,721 frames. 
As shown in Table~\ref{tab:coesot_results}, our method outperforms other SOTA trackers on multiple metrics, with SR and PR reaching 68.8 and 82.9, respectively. 
Among these SOTA trackers, ViPT~\cite{zhu2023visual}, SDSTrack~\cite{hou2024sdstrack}, and UnTrack~\cite{wu2024single} all utilize OSTrack pretrained weights for initialization. Compared with these trackers, our method achieves significant improvements in SR and PR metrics.

\noindent $\bullet$ \textbf{Results on VisEvent Dataset.}
\begin{table}
\centering
\small
\caption{Experimental results on VisEvent dataset.} 
\label{tab:visevent_results}
\begin{tabular}{l|c|cc}
\toprule
\textbf{Trackers} & \textbf{Source} &\textbf{SR}  &\textbf{PR} \\
\hline
\textbf{01. ATOM~\cite{danelljan2019atom}} & CVPR19 & 41.2 & 60.8 \\
\textbf{02. DiMP50~\cite{bhat2019learning}} & ICCV19 & 45.1 & 66.1 \\
\textbf{03. SiamCAR~\cite{guo2020siamcar}} & CVPR20 & 42.0 & 59.9 \\
\textbf{04. PrDiMP50~\cite{danelljan2020probabilistic}} & CVPR20 & 45.3 & 64.4 \\
\textbf{05. SiamR-CNN~\cite{voigtlaender2020siam}} & CVPR20 & 49.9 & 65.9 \\
\textbf{06. Stark~\cite{yan2021learning}} & ICCV21 & 44.6 & 61.2 \\
\textbf{07. TransT~\cite{chen2021transformer}} & ECCV22 & 47.4 & 65.0 \\
\textbf{08. OSTrack~\cite{ye2022joint}} & ECCV22 & 53.4 & 69.5 \\
\textbf{09. MDNet~\cite{wang2023visevent}} & TCYB23 & 42.6 & 66.1 \\
\textbf{10. ViPT~\cite{zhu2023visual}} & CVPR23 & 59.2 & 75.8 \\
\textbf{11. SDSTrack~\cite{hou2024sdstrack}} & CVPR24 & \textbf{59.7} & \textbf{76.7} \\
\textbf{12. UnTrack~\cite{wu2024single}} & CVPR24 & \textbf{59.7} & 76.3 \\
\hline
\textbf{13. Ours} & - & 58.8 & 76.6 \\
\bottomrule
\end{tabular} 
\end{table}
VisEvent~\cite{wang2023visevent} contains 820 video sequences for a total of 371,127 frames. 
As shown in Table~\ref{tab:visevent_results}, our method achieves SR of 58.8 and PR of 76.6, respectively.
Compared with SOTA trackers, our method has similar PR but slightly lower SR. 
We attribute the slightly lower SR to the relatively short average sequence length in VisEvent, yet our method remains competitive with SOTA trackers.

\begin{table*}
\centering
\small
\caption{Performance of different weights for each dataset.} 
\label{tab:ablation_weights}
\begin{tabular}{c|ccc|ccc|ccc|ccc}
\toprule
\multirow{2}{*}{\textbf{Weights}} & \multicolumn{3}{c|}{\textbf{COESOT}} & \multicolumn{3}{c|}{\textbf{FE108}} & \multicolumn{3}{c|}{\textbf{FELT}} & \multicolumn{3}{c}{\textbf{VisEvent}}\\
\cline{2-13}
& \textbf{SR} & \textbf{PR} & \textbf{NPR} & \textbf{SR} & \textbf{PR} & \textbf{NPR} & \textbf{SR} & \textbf{PR} & \textbf{NPR} & \textbf{SR} & \textbf{PR} & \textbf{NPR}\\
\hline 
MAE & 64.8 & 79.4 & 78.2 & 63.0 & 94.4 & 68.1 & 53.5 & 66.9 & 64.6 & 55.1 & 73.1 & 68.1 \\ 
OSTrack & 68.8 & 82.9 & 81.8 & 65.6 & 95.9 & 71.8 & 54.8 & 67.9 & 65.7 & 58.8 & 76.6 & 71.9 \\ 
\bottomrule
\end{tabular} 
\end{table*}

\subsection{Ablation Study} 
\noindent $\bullet$ \textbf{Analysis on Different Weights.} 
As shown in Table~\ref{tab:ablation_weights}, we comprehensively evaluate our method using two pretrained weights (MAE and OSTrack).
Compared to MAE pretrained weights, OSTrack pretrained weights lead to remarkable improvements. 
Across the FE108, FELT, COESOT, and VisEvent datasets, SR, PR, and NPR all increase. 
This consistent improvement across all four datasets highlights the importance of strong RGB pretrained weights.
Fine-tuned on the pretrained OSTrack, our model acquires rich visual representations and tracking behaviors, facilitating effective knowledge transfer and enabling fast adaptation across different datasets.

\noindent $\bullet$ \textbf{Effectiveness of Proposed Components.} 
\begin{table}
\centering
\small
\caption{Component analysis on COESOT and FE108.} 
\label{tab:ablation_components}
\setlength{\tabcolsep}{1mm} 
\begin{tabular}{c|cc|ccc|ccc}
\toprule
\multirow{2}{*}{\textbf{\#}} & \multirow{2}{*}{\textbf{ATU}} & \multirow{2}{*}{\textbf{AMAH}}  & \multicolumn{3}{c|}{\textbf{COESOT}} & \multicolumn{3}{c}{\textbf{FE108}} \\
\cline{4-9}
& & & \textbf{SR} & \textbf{PR} & \textbf{NPR} & \textbf{SR} & \textbf{PR} & \textbf{NPR} \\
\hline 
1 & \null & \null  & 63.2 & 77.1 & 76.1 & 60.4 & 91.6 & 65.0 \\ 
2 & \ding{51} & \null  & 64.1 & 78.7 & 77.5 & 61.0 & 92.3 & 65.3 \\ 
3 & \null & \ding{51}  & 63.9 & 78.4 & 77.1 & 62.0 & 93.2 & 67.6 \\
4 & \ding{51} & \ding{51} & \textbf{64.8} & \textbf{79.4} & \textbf{78.2} & \textbf{63.0} & \textbf{94.4} & \textbf{68.1}  \\
\bottomrule
\end{tabular} 
\end{table}
We conduct ablation experiments on each module using MAE pretrained weights. 
Table~\ref{tab:ablation_components} shows the effectiveness of Associative Template Update (ATU) and Associative Memory Augmented Hopfield (AMAH). ATU improves performance on both datasets by associatively updating templates to adapt to target changes, while AMAH also enhances performance by boosting dual-modal fusion in the backbone.
When both components are integrated, tracking performance peaks.

\noindent $\bullet$ \textbf{Analysis on Associative Template Update.} 

\begin{table}
\centering
\small
\caption{Analysis of ATU on COESOT.} 
\label{tab:ablation_ATU}
\begin{tabular}{c|c|ccc}
\toprule
\textbf{\#} & \textbf{\# Settings} & \textbf{SR} & \textbf{PR} & \textbf{NPR} \\
\hline 
\multirow{2}{*}{\makecell{Hopfield}} 
& \ding{55} & 63.9 & 78.0 & 77.3 \\ 
& \ding{51} & \textbf{64.8} & \textbf{79.4} & \textbf{78.2} \\ 
\hline 
\multirow{4}{*}{\makecell{$\bm{\beta}$}}
& 0.25 & 63.9 & 78.6 & 77.5 \\ 
& 1.0 & 64.2 & 78.5 & 77.4 \\ 
& \textbf{4.0} & \textbf{64.8} & \textbf{79.4} & \textbf{78.2} \\ 
& 16.0 & 64.5 & 78.9 & 78.0 \\ 
\hline 
\multirow{6}{*}{\makecell{LT/ST}} 
& 3 / 6 & 64.6 & 79.1 & 77.9 \\ 
& \textbf{5 / 10} & \textbf{64.8} & \textbf{79.4} & \textbf{78.2} \\ 
& 7 / 14 & 64.5 & 79.1 & 77.9 \\ 
& 14 / 7 & 64.6 & 79.2 & 78.0 \\ 
& 10 / 5 & 64.6 & 79.1 & 78.0 \\ 
& 6 / 3 & 64.5 & 79.0 & 77.8 \\ 
\hline 
\multirow{5}{*}{\makecell{RI/UI}} 
& 5 / 10 & 64.7 & 79.3 & 78.2 \\ 
& 5 / 20 & 64.6 & 79.1 & 77.9 \\ 
& \textbf{10 / 20} & \textbf{64.8} & \textbf{79.4} & \textbf{78.2} \\ 
& 20 / 20 & 64.7 & 79.2 & 78.0 \\ 
& 20 / 40 & 64.5 & 79.1 & 77.8 \\  
\hline 
\multirow{3}{*}{\makecell{RT}} 
& ST \& ST & 64.6 & 79.2 & 78.0 \\ 
& LT \& LT & 64.7 & 79.3 & 78.1 \\ 
& \textbf{ST \& LT} & \textbf{64.8} & \textbf{79.4} & \textbf{78.2} \\ 
\bottomrule
\end{tabular} 
\end{table}
As shown in Table~\ref{tab:ablation_ATU}, we analyze parameters of associative memory-based dynamic template updates.
First, we compare using and not using the Hopfield layer to enhance dynamic templates. The results show significant improvement with the Hopfield layer.
Then, we analyze multiple template update parameters: $\beta$ in the Hopfield layer, short- and long-term template memory lengths (ST and LT), template resampling and update intervals (RI and UI), and the resampling choice between short- and long-term template memory (RT).
Finally, we set $\beta$ to 4.0, the lengths of LT and ST are 5 and 10, respectively, the lengths of RI and UI are 10 and 20, respectively, and the resampling process selects dynamic templates from ST and LT, respectively. Our tracker then achieves the best performance in tracking.



\noindent $\bullet$ \textbf{Analysis on Associative Memory Augmented Hopfield.} 
\begin{table}[tb]
\centering
\small
\caption{Analysis of AMAH on COESOT.} 
\label{tab:ablation_AMAH}
\begin{tabular}{c|c|ccc}
\toprule
\textbf{\#} & \textbf{\# Settings} & \textbf{SR} & \textbf{PR} & \textbf{NPR} \\
\hline
\multirow{4}{*}{\makecell{$\bm{\beta}$}}
& 0.1 & 63.8 & 78.2 & 77.1 \\ 
& \textbf{0.25} & \textbf{63.9} & \textbf{78.4} & \textbf{77.1} \\ 
& 1.0 & 63.5 & 77.6 & 76.6  \\ 
& 4.0 & 62.6 & 76.7 & 75.9 \\ 
\hline
\multirow{3}{*}{\makecell{Layers}} 
& $[3,7,11]$ & 63.4 & 77.5 & 76.6 \\ 
& $[5,8,11]^{1}$ & 63.6 & 77.9 & 76.7 \\ 
& \bm{$[5,8,11]^{2}$} & \textbf{63.9} & \textbf{78.4} & \textbf{77.1} \\ 
\hline 
\multirow{5}{*}{\makecell{Model}} 
& Baseline & 63.2 & 77.1 & 76.1 \\
& MLP & 63.2 & 77.3 & 76.2 \\
& HopField Pooling & 63.2 & 77.5 & 76.7 \\
& HopField & 63.5 & 77.4 & 76.8  \\
& \textbf{Cross HopField} & \textbf{63.9} & \textbf{78.4} & \textbf{77.1}  \\
\bottomrule
\end{tabular} 
\end{table}
Table~\ref{tab:ablation_AMAH} shows the analysis of Associative Memory Augmented Hopfield.
To demonstrate the effectiveness of cross-modal retrieval of the Hopfield layer, we compare MLP, Hopfield Pooling, and the Hopfield Layer without cross-modal retrieval, keeping the selected layer consistent across experiments. The results show that our Hopfield layer's cross-modal retrieval performs better.
As shown in Table~\ref{tab:ablation_AMAH}, we also analyze and compare cross-modal Hopfield retrieval layer hyperparameters.
Setting $\beta$ to 0.25, which superposes multiple stored patterns as the result, yields the best performance and enables multi-layer contextual feature integration.
We then analyze the detailed selection of Hopfield layers and preceding layers as stored patterns.
The selection of stored patterns and preceding layer features in the Hopfield layer is crucial. We compare three schemes: 
\textit{1).} $[3,7,11]$ (\(\mathcal{L}=\{3,7,11\}\), \(\mathcal{P}_3=\{0,1,2\}\), \(\mathcal{P}_7=\{4,5,6\}\), \(\mathcal{P}_{11}=\{8,9,10\}\)).
\textit{2).} $[5,8,11]^{1}$ (\(\mathcal{L}=\{5,8,11\}\), \(\mathcal{P}_5=\{1,2,3\}\), \(\mathcal{P}_8=\{4,5,6\}\), \(\mathcal{P}_{11}=\{7,8,9\}\)).
\textit{3).} $[5,8,11]^{2}$ (\(\mathcal{L}=\{5,8,11\}\), \(\mathcal{P}_5=\{1,3\}\), \(\mathcal{P}_8=\{4,6\}\), \(\mathcal{P}_{11}=\{7,9\}\)).
Finally, we choose $[5,8,11]^{2}$ as the setting of our Hopfield retrieval layer. The other two settings also improve the performance compared to the baseline.

\begin{figure*}[!tb]
\centering
\includegraphics[width=\linewidth]{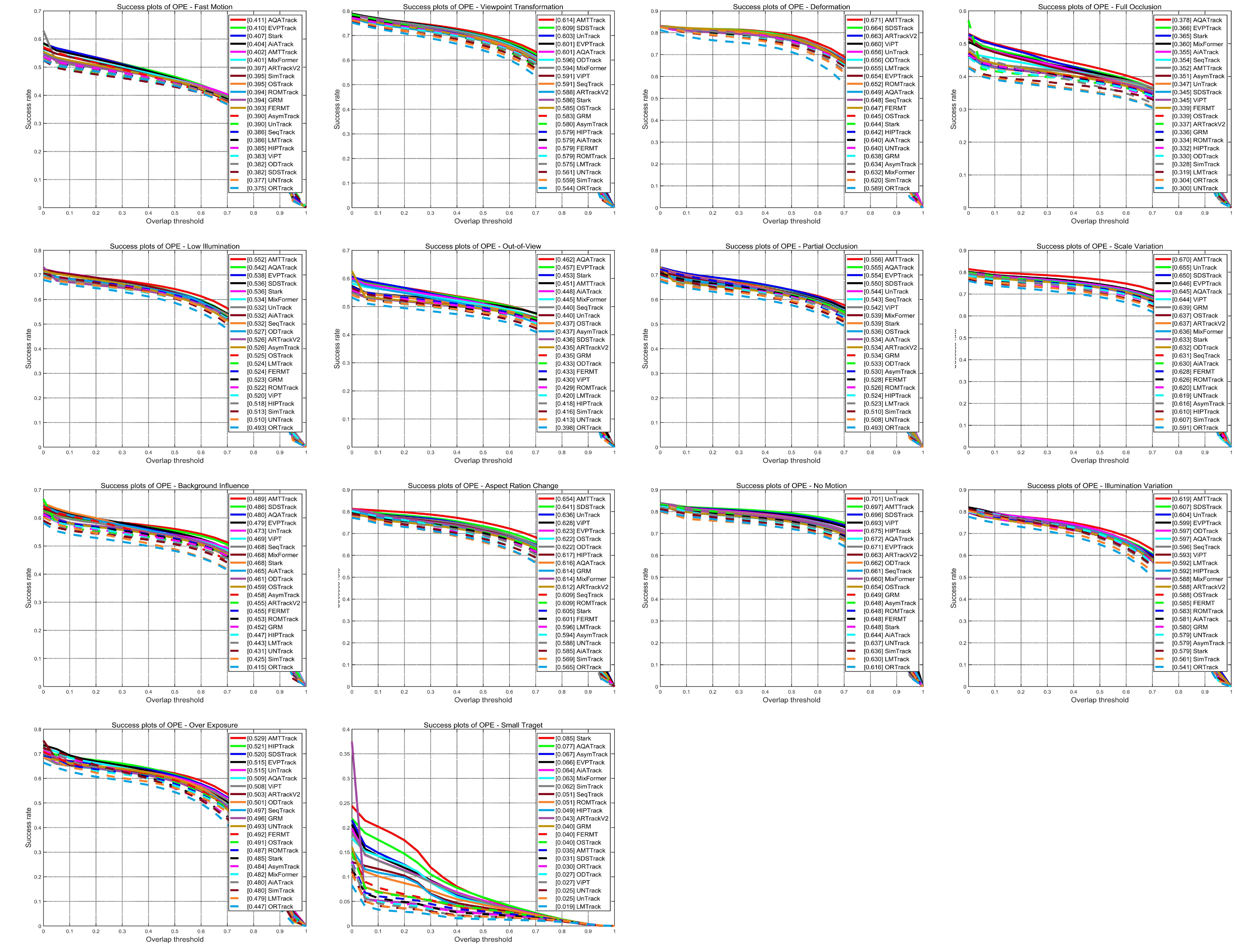}
\caption{Tracking results (SR) under 14 challenging factors.}
\label{fig:felt_attributesResults}
\end{figure*}

\noindent $\bullet$ \textbf{Tracking Results Under Challenging Scenarios.} 
In this paper, as shown in Table~\ref{tab:felt_attributes}, we define 14 attributes for the evaluation of visual trackers when facing challenging factors. 
As shown in Fig.~\ref{fig:felt_attributesResults}, our trackers can achieve good performance in most challenging situations including VT (View Transformation), DEF (Deformation), SV (Scale Variation) and OE (Over Exposure).




\subsection{Visualization} 
\begin{figure*}
\centering
\includegraphics[width=\linewidth]{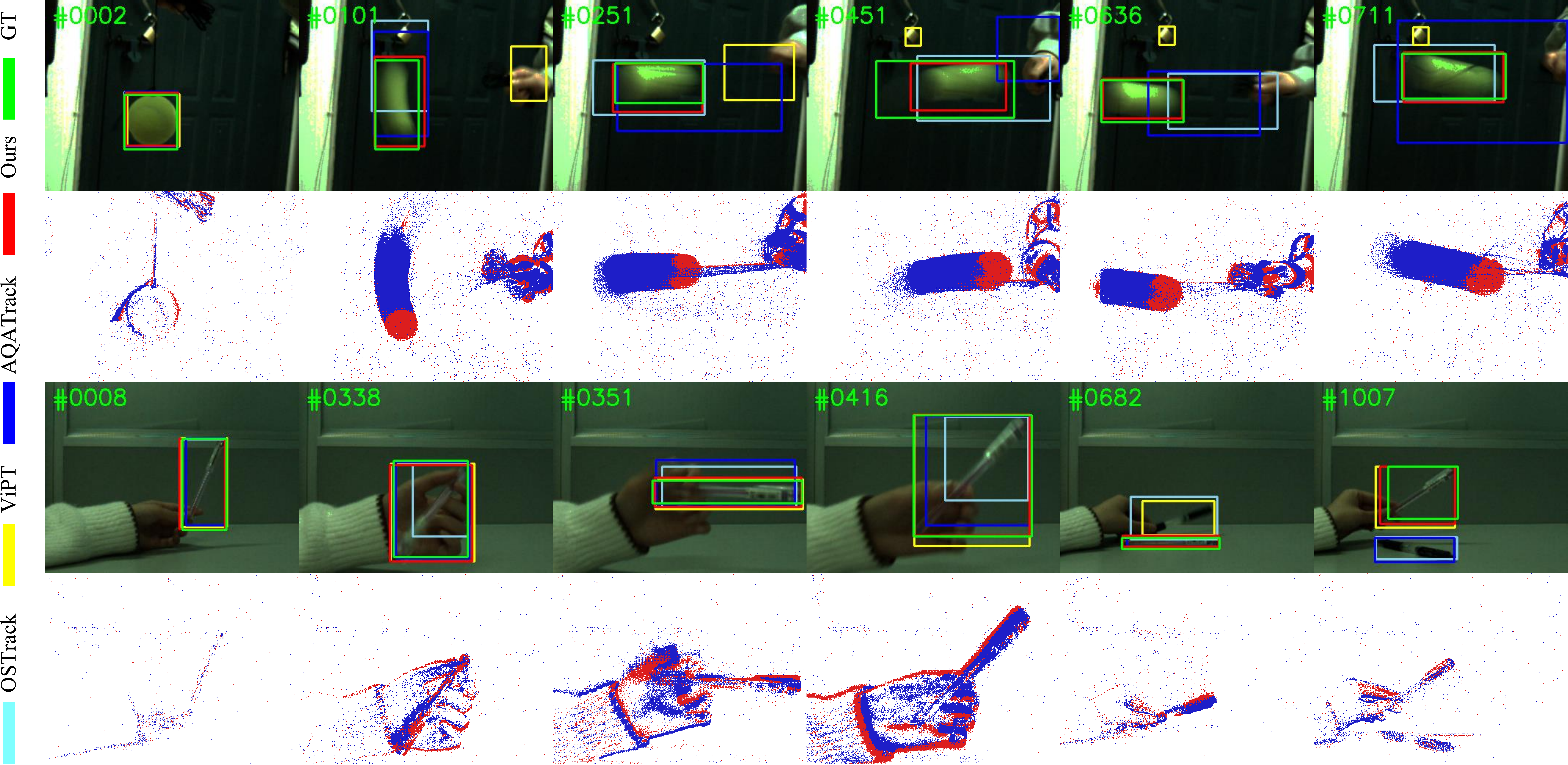}
\caption{Tracking results of our AMTTrack and other trackers.} 
\label{fig:trackingResults}
\end{figure*}

Beyond the quantitative analysis, we also provide a visualization to help readers better understand our tracking algorithm.
Fig.~\ref{fig:trackingResults} demonstrates the robust performance of our tracker against other SOTA visual trackers, confirming its effectiveness in RGB-Event long-term visual tracking.

\subsection{Limitation Analysis}  
While our method performs well on multiple datasets, some metrics can be optimized. For instance, on the FELT dataset, our PR is outperformed by temporal model-based trackers. In future work, we will leverage frame-to-frame temporal cues or explore the dense temporal resolution within event data to enhance long-term tracking.

\section{Conclusion}
In this work, we propose a new long-term and large-scale frame-event single object tracking dataset, termed FELT. It contains 1,044 videos and 1,949,680 RGB frames and event stream pairs and is also the largest frame-event tracking dataset to date.
To build a comprehensive benchmark, we also retrain and evaluate multiple baseline trackers on our newly proposed FELT SOT dataset, which will provide a better platform for future works to make comparisons. 
In addition, we propose AMTTrack, a novel associative memory-augmented transformer-based tracker. It utilizes modern Hopfield layers to jointly optimize multi-modal feature extraction and representation within the backbone transformer network while refining dynamic templates, thereby enhancing the quality of multi-modal long-term tracking.
Extensive experiments on multiple RGB-Event based tracking datasets fully validate the effectiveness of our proposed model.

\clearpage
{
    \small
    \bibliographystyle{ieeenat_fullname}
    \bibliography{main}
}


\end{document}